# The Frankfurt Latin Lexicon

## From Morphological Expansion and Word Embeddings to SemioGraphs

Alexander Mehler, Bernhard Jussen, Tim Geelhaar, Alexander Henlein, Giuseppe Abrami, Daniel Baumartz, Tolga Uslu and Wahed Hemati
Goethe Universität Frankfurt am Main
Robert-Mayer-Straße 10
D-60325 Frankfurt am Main
mehler@em.uni-frankfurt.de

**Abstract:** In this article we present the *Frankfurt Latin Lexicon* (FLL), a lexical resource for Medieval Latin that is used both for the lemmatization of Latin texts and for the post-editing of lemmatizations. We describe recent advances in the development of lemmatizers and test them against the Capitularies corpus (comprising Frankish royal edicts, mid-6th to mid-9th century), a corpus created as a reference for processing Medieval Latin. We also consider the post-correction of lemmatizations using a limited crowdsourcing process aimed at continuous review and updating of the FLL. Starting from the texts resulting from this lemmatization process, we describe the extension of the FLL by means of word embeddings, whose interactive traversing by means of SemioGraphs completes the digital enhanced hermeneutic circle. In this way, the article argues for a more comprehensive understanding of lemmatization, encompassing classical machine learning as well as intellectual post-corrections and, in particular, human computation in the form of interpretation processes based on graph representations of the underlying lexical resources.

### 1. *Introduction*

Regarding lexical resources for *Natural Language Processing* (NLP) of historical languages such as Latin, three paradigms can be roughly distinguished: (1) morphologically enriched lexica or dictionaries such as the *Frankfurt Latin Lexicon* (FLL) to be presented here, which use, for example, rules of morphological expansion to generate inflected forms from lemmas collected from web and other resources, (2) wordnets such as the famous WordNet (Miller, 1995), which as a terminological ontology (Sowa, 2000) distinguishes (wordforms as search terms of) lemmata from synsets and their sense relations, and (3) word embeddings (Komninos and Manandhar, 2016; Levy and Goldberg, 2014; Ling et al., 2015; Mikolov et al., 2013) which address the statistical modeling of syntagmatic (contiguity) and paradigmatic (similarity) associations of lexical units (Halliday and Hasan, 1976; Jakobson, 1971; Miller and Charles, 1991; Raible, 1981).[1] Ideally, for a historical language such as Latin, there exists an integrated system of resources of these kinds in a sufficiently deep state of development: by using such a resource, a professional user or NLP system is extensively informed about the lexical units of the target language on different levels of lexical resolution (including wordforms, lemmata, super-lemmata, lexeme groups etc.), about their morpho-syntactic and semantic features as well as about their various sense relations and unsystematic associations. Regarding the example of lemmatization, such a resource would support both the initial automatic lemmatization and its intellectual post-correction, which in turn would be the starting point for the post-correction or further development of this resource, thereby closing the digitally enhanced hermeneutic circle. However, the example of the

[1] In the case of modern languages, a fourth paradigm would be given by knowledge graphs derived, for example, from Wikidata or Wikipedia.

Latin WordNet (Minozzi 2017) already shows that the components of such an integrated resource are still out of reach for this historical language (as explained and analyzed in Franzini et al., 2019). The same applies to input-intensive word embeddings, which require large amounts of text data, but which are not yet freely available for Latin (see, however, UDify (Kondratyuk and Straka, 2019) as an example of an approach that seems to circumvent this limitation – cf. Section 4). Last but not least, voluntarily created lexical information systems such as Wiktionary, which aim to integrate wordnet-related information with dictionary information, suffer not only from a lack of scope, but also from multiple sources of information biases (cf. Mehler et al. 2017). Therefore, it remains a challenge to provide not only one of the three types of resources (dictionary, wordnet, embeddings) for Latin in sufficient quantity, but even more to do so for at least two of these types – in an integrated manner. This article wants to take a step in this direction. That is, we present the FLL as a kind of Latin dictionary that distinguishes lexical units at the level of word forms, syntactic words[2], lemmata, and superlemmata, provides rich grammatical information for syntactic words, serves as a resource for the post-correction of automatic lemmatization, provides a word-for-word monitoring of the lemmatization status of each text, which is particularly easy to read for non-IT philologists, and supports the computation of word embeddings at various levels of lexical resolution. We also show how these embeddings can be presented as interactive graphs to encourage the correction and further development of the underlying resources.

The article is organized as follows: Section 2 outlines the structure of the FLL and quantifies the extent of its overgeneration or, conversely, its lack of coverage. Section 3 deals with the post-processing of the lemmatization of Latin texts with the help of the FLL, while Section 4 compares the current progress of lemmatizers for Latin. Subsequently, Section 5 deals with the derivation of genre-sensitive word embeddings for Latin and their visualization by means of interactive SemioGraphs. These visualizations are then used in Section 6 to conduct three case studies in computational historical semantics, which ultimately combine lemmatization and the evaluation of word embedding graphs with the underlying FLL. In this way, we will speak of a *digitally enhanced hermeneutic circle* implemented through the NLP pipeline for Latin, as presented in this article. Finally, Section 7 draws conclusions and gives an outlook on future work.

## 2. *From Superlemmata to Syntactic Words*

The *Frankfurt Latin Lexicon*[3] (FLL) is a morphological lexicon, currently for Medieval Latin, that is Latin between 400 and 1500 CE. Its main purpose is to support the automatic lemmatization of Latin texts with the *Text-technology Lab* Latin Tagger (TTLab Tagger) (Gleim et al., 2019; cf. Stoeckel 2020), which is available through the TextImager[4] (Hemati et al., 2016), the eHumanities Desktop[5] (Gleim et al., 2012) and GitHub[6]. It was created starting in 2009 (cf. Mehler et al., 2011; see also Jussen et al. 2007) by extracting and collecting lemmata from various web-based resources. This in-cludes[7] the AGFL Grammar Work Lab[8] (Koster and Verbruggen, 2002), the Latin morphological analyzer LemLat (Passarotti, 2004), the Perseus Digital Library (Crane, 1996), William Whitaker's Words[9], the Index Thomisticus[10] (Passarotti and Dell'Orletta, 2010), Ramminger's Neulateinische Wortliste[11], the Latin Wiktionary[12], Latin training data of the Tree Tagger (Schmid, 1994), the so-called Najock Thesaurus[13] , and other resources from cooperating projects like "Nomen et Gens" that provide several thousands of Latin personal names[14]. Since then, the FLL has grown continuously through the lemmatization of Latin Texts.[15]

[2] Syntactic words are signs in the sense of structuralism (Saussure 1916): they include an expression plane (called *wordform*) and a content plane. The content plane of syntactic words is usually represented by an attribute value-structure that collects grammatical features such as *case* and *numerus* in the case of nouns or *tempus* and *genus verbi* in the case of verbs. Thus, the same wordform may be mapped to different syntactic words (as, for example, *house* in "*Your house*$_1$ *is next to her house*$_2$" in which the tokens *house*$_1$ and *house*$_2$ manifest the same wordform but two different syntactic words distinguished by case.

[3] https://www.comphistsem.org/70.html/

[4] https://textimager.hucompute.org/

[5] https://hudesktop.hucompute.org/

[6] https://github.com/texttechnologylab

| PoS | Superlemma | Lemma | Syntactic Word | Description |
|---|---|---|---|---|
| ADJ | 21,870 | 26,070 | 3,337,028 | adjective |
| ADV | 9,682 | 11,163 | 42,864 | adverb |
| AP | 86 | 117 | 482 | preposition |
| CON | 101 | 140 | 519 | conjunction |
| DIST | 46 | 49 | 1,321 | distributive number |
| FM | 76 | 109 | 2,343 | Foreign Material |
| ITJ | 112 | 115 | 254 | interjection |
| NE | 5,843 | 6,649 | 114,757 | named entity |
| NN | 35,433 | 45,383 | 745,345 | common noun |
| NP | 26,741 | 29,657 | 247,911 | personal name |
| NUM | 101 | 143 | 3,140 | number |
| ORD | 131 | 194 | 4,871 | ordinal number |
| PRO | 125 | 172 | 8,041 | pronoun |
| PTC | 12 | 17 | 38 | particle |
| V | 9,081 | 13,164 | 5,135,824 | verb |
| XY | 114 | 117 | 718 | unknown |
| Sum | 109,554 | 133,259 | 9,645,456 | |

Table 1. *Statistics of the FLL (release as of May, 2019): superlemmas, lemmas and syntactic words are listed together with their numbers and differentiated by 15 parts of speech, supplemented by a class of words (denoted by XY), which collects unknown cases.*

The entries of the FLL are structured according to a four-level model consisting of wordforms, syntactic words (mapping wordforms onto vectors of grammatical features), lemmata and superlemmata. The introduction of the superlemma level was particularly important for preserving the considerable orthographical richness of Medieval Latin as a historical language. This approach is analogous to the Wiktionary model of lexical units, but in contrast to Wiktionaries and above all wordnets (such as the Latin WordNet – cf. Franzini et al. 2019) it lacks lexical-semantic relations (see Section 1).

The superlemma provides the normalized spelling of a lemma so that on the lemma level different spellings can be kept. The FLL currently contains 116,297 superlemmata and 133,691 subordinated lemmata. These lemmata have been expanded morphologically according to the standard grammar of Classical Latin as described in Menge (2009) and Rubenbauer et al. (2009) so that the FLL now has 9,663,808 syntactic words (see Table 1 for earlier statistics of the FLL).[16] In the future, the Superlemma-ID will be used to connect the FLL with other lexical resources on the web (and also with resources provided by traditional long-term institutions for Latin lexicography), so that morphological information will be available together with reading aids. The lexicon could also work with a fourth (*lexeme group*) and a fifth level (*synset*) to bundle superlemmata of different POS that share the same root, or to map semantic relations. But this is future work.

[7] For the following list see Mehler et al. (2015); see also vor der Brück and Mehler (2016) for more information about the FLL. The presentation of the FLL in this article is a correction of its earlier presentation in Mehler et al. (2015), which contained much higher amounts of overgeneration.
[8] Apparently, this resource no longer exists.
[9] http://archives.nd.edu/words.html; today http://www.latin-dictionary.net/.
[10] http://www.corpusthomisticum.org/it/index.age
[11] http://www.neulatein.de/
[12] https://la.wiktionary.org/wiki/Victionarium:Pagina_prima
[13] This data was provided by Michael Trauth, Trier University.
[14] http://www.neg.uni-tuebingen.de/
[15] This concerns mainly texts provided by the project *Corpus Corporum* of Philipp Roelli in Zurich (http://www.mlat.uzh.ch/MLS/), the *Monumenta Germaniae Historica* (https://www.dmgh.de/) and the *Institut de recherches d'histoire des textes* (IRHT; https://www.irht.cnrs.fr/).

The FLL can map multi-word units, which makes it easier to record proper names such as Colonia Agrippina. However, the four-step model currently meets the requirements of lemmatization. The lexicon is managed via the *Lexicon Browser* of the eHumanities Desktop, which has been especially adapted for humanities scholars without programming skills. Entries can be created, changed, merged, reorganized or deleted by authorized users. The so-called *Extension Tool* then creates all inflected forms for newly entered lemmata. Only basic information is needed to identify the right declination or conjugation for the new token. All changes are documented by naming the authors and timestamps. Additional columns allow descriptions to be added to the entries or to show if an entry has been double-checked. This procedure was developed together with various third-party funded humanities projects that have based their philological and linguistic research on their work with the FFL.[17]

| Attribute | Value |
|---|---|
| All texts | 111,515 |
| All tokens | 185,808,777 |
| Tokens mapped to the FLL | 180,535,369 (97.16%) |
| Tokens unassigned | 5,273,408 |
| All superlemmata in the FLL | 109,554 |
| Superlemmata used | 83,780 (76.47%) |
| All lemmata in the FLL | 133,259 |
| Lemmata used | 102,728 (77.09%) |
| All syntactic words | 9,645,456 |
| Syntactic words used | 871,452 (9.04%) |

Table 2. *On overgeneration and underrepresentation as induced by the FLL (release as of May, 2019).*

[16] Apart from some exceptions like the oblique case, the grammar rules did not alter between Classical and Medieval Latin; cf. (Menge, 2009; Rubenbauer et al., 2009).
[17] See below.

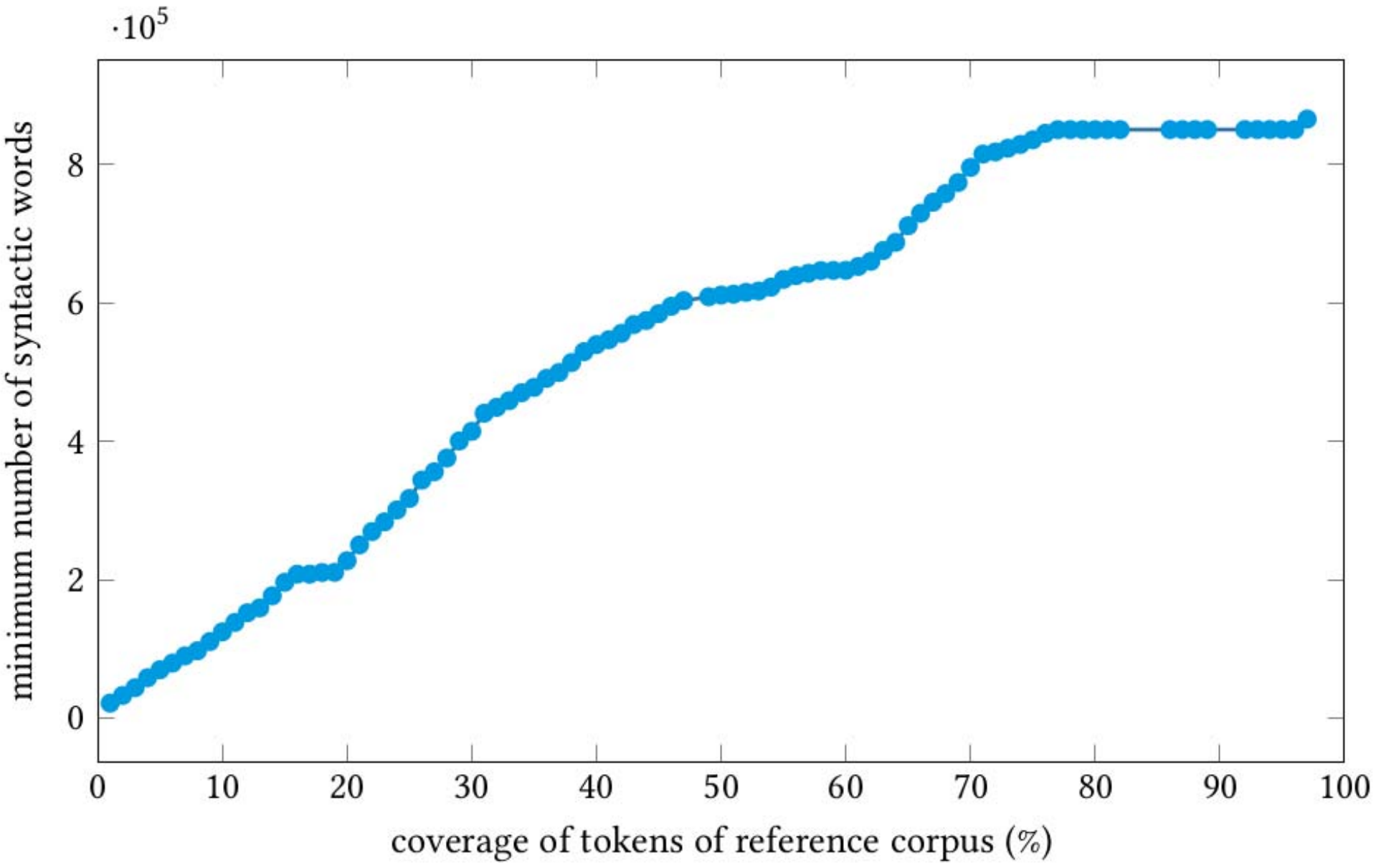

Figure 1. *The minimum number of syntactic words of the FLL (release as of May, 2019) (y-axis) sufficient to cover the corresponding percentage of tokens of the reference corpus.*

An objection to the method of morphological expansion, and thus to lexica of the type of the FLL, is to say that it is prone to overgeneration. To calculate this, we used as a reference corpus a repository of Latin texts including *Migne Patrologia Latina* (MPL), substantial parts of the *Monumenta Germaniae Historica* (MGH) and other repositories[18] to ask for the number of syntactic words of the FLL that our lemmatization finds manifested in this repository. This is shown in Table 2. Indeed, only 9% of the syntactic words of the FLL are found in this "text repository". Figure 1 shows how this coverage grows with the percentage of tokens of the reference corpus covered by the FLL. It demonstrates almost a linear trend – except for the rightmost part of the distribution: the number of syntactic words in the FLL that are observed in the corpus grows linearly with the size of this corpus. The coverages are much higher on the level of lemmata (77.09%) and superlemmata (76.47%). On the other hand, the underrepresentation, that is, the number of tokens within our reference corpus that are unknown from the point of view of the FLL, is remarkably low (2.84%). Such a rate of coverage did not seem to be possible in the early days of the FLL: in fact, it was the morphological expansion that made it an extensive lexicon that allows such rates, so that with each lemmatized token the associated grammatical features can be linked. Furthermore, the corpus of Medieval Latin texts within the eHumanities Desktop[19] is continuously extended, with each syntactic word being identified by a corresponding corpus frequency. These frequencies allow the subsequent filtering of supposedly overgenerated words for downstream tasks or even for information retrieval by users.

[18] For the *Monumenta Germaniae Historica* (MGH) see the openMGH repository (http://www.mgh.de/dmgh/openmgh/); *Migne Patrologia Latina* (MPL) is available from the *Corpus Corporum* website (http://www.mlat.uzh.ch/MLS/); in addition the repository includes the *Roman Law Library* (https://droitromain.univ-grenoble-alpes.fr/), the corpus of *Cluny Charters* (http://www.cbma-project.eu/), parts from the *Latin Library* (http://thelatinlibrary.com/) and from the *Central European Medieval Texts Series* (http://ceupress.com/series/central-european-medieval-texts).

[19] It currently manages 112,657 Latin documents of different sizes (release as of November 2019).

## 3. *Crowdsourcing the FLL*

The FLL grows as new texts are uploaded into the text database of HSCM[20] and lemmatized. The result of automatic lemmatization is checked mainly by Latin philologists during the post-lemmatization process. They correct unfitting assignments between text and lexicon or create new entries in the FLL to close gaps in the lexicon and in the lemmatization. To this end, human editors can use the so-called *Lemmatization Editor* of the eHumanities Desktop. It presents lemmatized text in a color code and a statistical overview indicating the state of lemmatization. The color code differentiates nine distinct levels of lemmatization. The code does not only lead the human editor to tokens that still need to be identified or disambiguated but also marks lemmatization results with different degrees of certainty. The expert then opens the so-called *Word-Link-Editor* for a token to check the tagger's choice that the expert can confirm or correct. Here, she or he can disambiguate the result if the tagger has not taken a decision. In very few cases – concerning mainly proper names, OCR mistakes or abbreviated word-forms – the color code displays the token in blue which means that no corresponding entry could be found in the FLL. In this case, the human editor can either correct the misspelling directly within the Lemmatization Editor or create a new entry. If necessary, the expert can leave the editor and open the Lexicon Browser to create a new superlemma and/or expand lemmata. All these actions influence the state of lemmatization directly which becomes visible through the changing color code.

Evidently, the manual post-lemmatization process may detect errors in the FLL. In this case, the expert must correct the corresponding entries, merge duplicates, or delete incorrect entries. Such errors are mainly due to the initial setup when information was taken from different sources or through overgeneration as a result of morphological expansion. But even human editors sometimes make mistakes. Therefore the lexicon offers the possibility to mark entries as *double-checked*. Since changes in the lexicon cannot simply be undone, changing the entries requires a high level of expertise, which is why the lemmatization and subsequent lexicon work is only carried out by trustworthy project members. As a result, this work is done using a limited crowdsourcing approach by assigning update rights to a limited number of experts using eHumanities Desktop's rights management tool (Gleim et al., 2012). In the project *Computational Historical Semantics*[21], linguists from the universities of Bielefeld, Regensburg and Tübingen, who worked directly with the TTLab and the Historical Seminar of the Goethe University, were allowed to update the lexicon. In addition, external partner projects from the Universities of Cologne, Freiburg and Mainz participate in the update process of the FLL after appropriate training.[22] As of May 2019, the percentage of lemmata created or modified by this procedure was 13.93% (18,565 lemmata).

---

[20] HSCM stands for *Historical Semantics Corpus Management*, a system for managing the Latin text database of the eHumanities Desktop.

[21] www.comphistsem.org

[22] This concerns e.g. the project HUMANIST (2017-2020 at the universities Darmstadt, Mainz and at the Hochschule Mainz; https://humanist.hs-mainz.de/, funded by the Federal Ministry of Education and Research). In this context, a current project at Johann Gutenberg University is establishing a digital version of the so-called various letters written by the eminent 6th century politician and philosopher Cassiodorus (d. ca. 585). Here, specialists are checking the automatic lemmatization provided by means of the FLL and produce completely disambiguated texts for their project's purposes. Particularly worth emphasizing is how they make their work transparent – see https://humanist.hs-mainz.de/projekt/inhaltlicher-projektkern/digitale-edition/; a Freiburg University based partner project, funded by the German Research Foundation, focused on high medieval feudal law and imperial charters (https://gepris.dfg.de/gepris/projekt/264932155); another partner project focusses on 6th to 9th century Frankish royal edicts, so called capitularies (https://capitularia.uni-koeln.de/, a long term project funded by the Union of the German Academies of Sciences and Humanities).

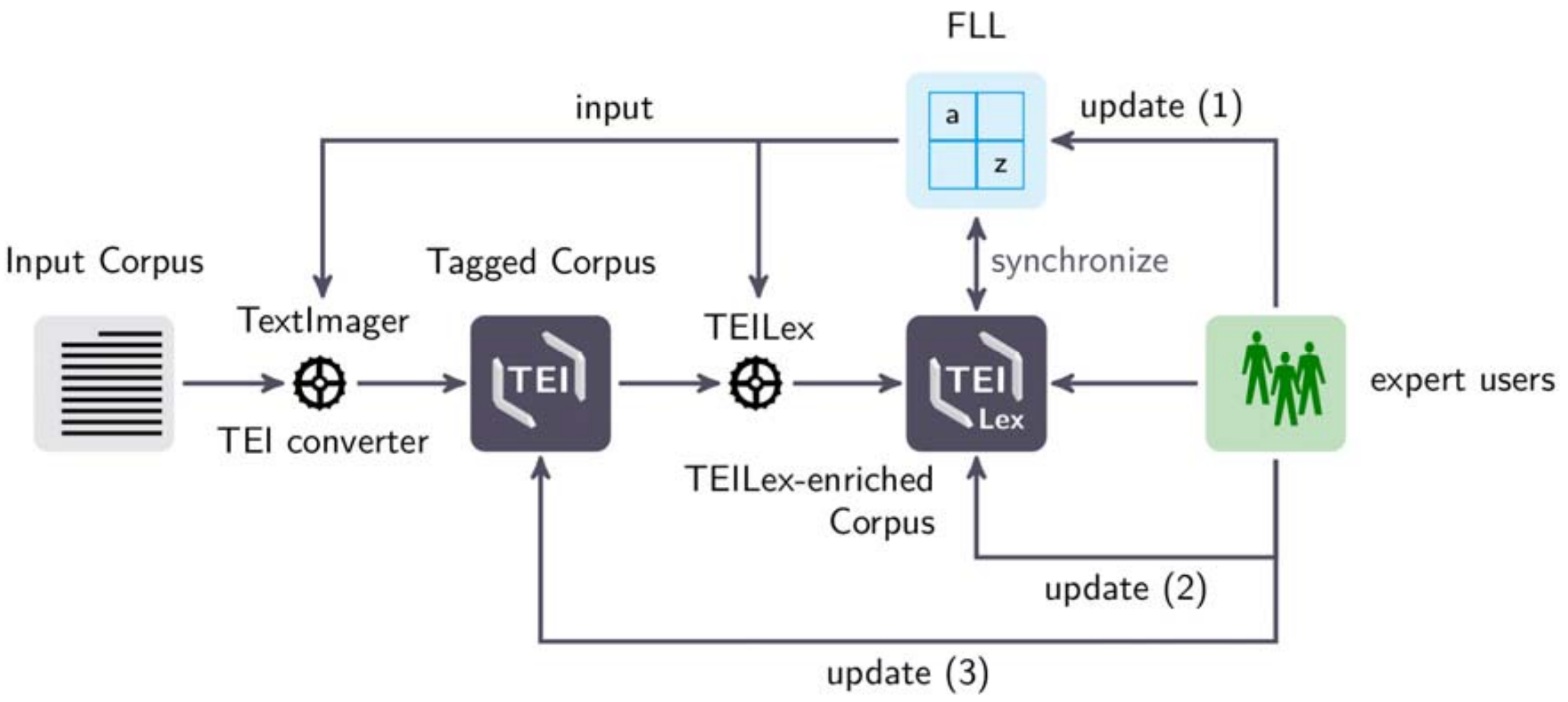

Figure 2. *Schematic depiction of TEILex' workflow.*

The synchronization between the lemmatized corpus and the FLL as induced by post-lemmatization and lexicon updates is managed by TEILex, a system for integrating lexica and text corpora, in which the tokens of a corpus are linked with corresponding lexicon entries in such a way that lexicon updates are immediately transferred to the linked corpora and vice versa. In this way, expert-based lexicon modeling becomes less dependent of indexing the underlying corpus. Figure 2 shows the corresponding workflow of TEILex: automatic text processing using TextImager (which generates XMI files) and subsequent TEI conversion generates a TEI-conform corpus that is indexed and synchronized with the FLL using TEILex. In this process, experts can change both the lexicon (see update (1) in Figure 2) and the tagged corpus (update (3)). However, in order to prevent the corpus from being re-indexed after each change, the synchronization of the TEILex corpus with the FLL allows the automatic execution of changes of the lexicon on the synchronized corpora. The TEILex index is then automatically revised without having to interrupt the use of the system. This procedure has greatly accelerated the post-correction process by protecting it from too many interruptions.

## 4. *Lemmatization in the Flux*

In Eger et al. (2016) and Gleim et al. (2019), we experimented with different lemmatizers on Latin datasets. Since the publication, thanks to the emergence of large transformer networks (e.g. BERT; Devlin et al., 2019), significantly better models have appeared, which prompted us to update our results. These transformer networks are large language models that are trained on even larger amounts of data and recognize and process syntactic and semantic relations (Clark et al., 2019). These attention-based networks with up to 24 layers and over 350 million parameters (sometimes even more; see Shoeybi et al., 2019) are trained on natural language texts (mostly Wikipedia and digital books) with the task of reconstructing deleted words by their context. These models can then be adapted to specific tasks on the basis of training data (Kondratyuk and Straka, 2019). BERT (Devlin et al., 2019) is still the most popular model, as it was pre-trained in 104 languages and is publicly available. In addition, the models are not too large, so that they can be (post-)trained for individual purposes without the need for special hardware. The most advanced lemmatization model currently available from this class of approaches is UDify (Kondratyuk and Straka, 2019), which is based on a multilingual BERT model. UDify has been fine-tuned on 124 treebanks in 75 languages and is capable of tagging universal POS[23], morphological features, lemmata, and dependency trees and also obtains acceptable results in unknown languages. Under these 124 treebanks are 3 in Latin: Proiel (Haug and Jøhndal, 2008), Perseus (Bamman and Crane, 2011) and ITTB (Cecchini et al., 2018; Passarotti and Dell'Orletta, 2010) with a total of 582,336 tokens. The Proiel treebank contains most of the Vulgate New Testament translations plus selections from Caesar and Cicero. ITTB (i.e. the Index Thomisticus Treebank) contains the complete work by Thomas Aquinas (1225–1274; Medieval Latin) and by 61 other authors related to Thomas, while Perseus contains a selection of passages from diverse authors like Augustus and Tacitus. All treebanks are annotated with the *Universal Dependencies*[24] (UD) Framework (Nivre et al., 2016). In order to adapt UDify, the texts are preprocessed with the help of BERT, whereby for each downstream task a separate classifier is trained with the help of the resulting BERT word vectors. The UDPipe model (Straka and Straková, 2017), on the other hand, is a pipeline system (i.e. a system that interconnects series of NLP tools) that does not rely on large transformer models, but is designed independently for each target language and therefore cannot (directly) exploit similarities between languages. As a consequence, 97 independent models were trained on 97 treebanks from 64 languages (Straka and Straková, 2017).

[23] For a recently published study on POS tagging in Latin see Stoeckel et al. (2020).

| Lemmatizer | Trainings Corpus | Proiel (Haug and Jøhndal, 2008) | Capitularies (Mehler et al., 2015) |
|---|---|---|---|
| LemmaGen (Juršic et al., 2010) | Capitularies (Mehler et al., 2015) | 81.39 (Gleim et al., 2019) | 95.64 (Gleim et al., 2019) |
| MarMoT (Müller et al., 2013) | Capitularies (Mehler et al., 2015) | 81.24 (Gleim et al., 2019) | 95.81 (Gleim et al., 2019) |
| LemmaTag (Kondratyuk et al., 2018) | Capitularies (Mehler et al., 2015) | 82.25 (Gleim et al., 2019) | 96.13 (Gleim et al., 2019) |
| LemmaGen (Juršic et al., 2010) | Proiel (Haug and Jøhndal, 2008) | 90.63 (Gleim et al., 2019) | 76.28 (Gleim et al., 2019) |
| MarMoT (Müller et al., 2013) | Proiel (Haug and Jøhndal, 2008) | 90.29 (Gleim et al., 2019) | 76.37 (Gleim et al., 2019) |
| LemmaTag (Kondratyuk et al., 2018) | Proiel (Haug and Jøhndal, 2008) | 81.85 (Gleim et al., 2019) | 49.61 (Gleim et al., 2019) |
| UDPipe (Straka and Straková, 2017) | ITTB (Passarotti and Dell'Orletta, 2010) | --- | 83.80 |
| UDPipe (Straka and Straková, 2017) | Perseus (Bamman and Crane, 2011) | --- | 78.87 |
| UDPipe (Straka and Straková, 2017) | Proiel (Haug and Jøhndal, 2008) | 96.32 (Kondratyuk and Straka, 2019) | 86.94 |
| UDify (Kondratyuk and Straka, 2019) | 124 treebanks | 91.79 (Kondratyuk and Straka, 2019) | 88.25 |

Table 3. *Results of lemmatizers trained on different data (rows) and evaluated on the Proiel corpus and our Capitularies (Frankish royal edicts, 6th to 9th c.) corpus (columns). Blue indicates best results of in-domain and red of out-domain lemmatization. The references behind the results refer to the paper in which they were published. Unreferenced scores indicate newly trained models.*

| Form | Gold | Predicted | Count |
|---|---|---|---|
| a | a | ab | 2020 |
| se | sui | se | 999 |
| quod | quod | qui | 892 |
| ac | ac | atque | 884 |
| sibi | sui | se | 371 |
| vero | vero | verus | 345 |
| seu | seu | sive | 342 |

Table 4. *Most frequent errors made by UDify on the Capitularies.*

We have tested both models on our corpus of Frankish royal edicts, the capitularies (Mehler et al., 2015), and on the Proiel corpus (Haug and Jøhndal, 2008) and compared the results with the taggers from our original work (Gleim et al. 2019). All results are listed in Table 3. First we concentrate on the results on the Proiel corpus. UDPipe achieves significantly better results on this dataset than UDify, although both were trained on this dataset. Generally with 96.32% a very good F1 score[25] is achieved by UDPipe. However, the dataset used by UDify for training also included the ITTB and Perseus data. It is not surprising that the tools that were trained on the Capitularies perform significantly worse on the Proiel data. The evaluation on the Capitularies, on the other hand, is more interesting since neither UDPipe nor UDify were trained on it. LemmaTag, which was also trained on the Capitularies (Gleim et al., 2019), reaches an F1 score of more than 96%. Taggers such as LemmaGen, MarMoT and LemmaTag on the other hand, which were only trained on Proiel, generalize much worse when being evaluated out-domain by means of the Capitularies; this can be an indicator of overfitting. UDify, which was trained on three Latin corpora and several other languages, generalizes much better: being evaluated out-domain by means of the Capitularies, it still reaches an F1 score of 88.25%. Among all taggers which were not trained on the Capitularies, UDify achieves the best results. This ability to generalize makes it a very interesting candidate for lemmatization. The results between the corpora may have been even stronger, but there are differences in annotation between them. This is particularly evident in the errors that UDify makes most often (as listed in Table 4). Just by fixing these errors by means of a simple post-processor, UDify's performance can be considerably improved.

[24] https://universaldependencies.org/

| | #Text | #Tokens |
|---|---|---|
| Reference corpus | 111,515 | 185,808,777 |
| Overall training corpus | 33,791 | 61,451,677 |
| Epistolographic texts | 844 | 16,406,556 |
| Legal texts | 31,461 | 12,097,990 |
| Liturgical texts | 252 | 2,667,784 |
| Narrative texts | 663 | 7,635,906 |
| Political texts | 31 | 3,197,879 |
| Theological texts | 494 | 18,305,475 |

Table 5. *Statistics of the corpora used for computing specialized embeddings.*

This analysis shows that in-domain lemmatization can be delegated to modern neural network models that appear to be largely independent of lexicons of the type of the FLL. Even those models of Gleim et al. (2019), which use the FLL, are outperformed by transformer-based models in the area of out-domain lemmatization. From the point of view of the manual post-lemmatization process, however, the reference to a lexicon remains indispensable when it comes to distinguishing between lemmata and superlemmata and correctly assigning them to incorrectly lemmatized tokens – a residual task that can probably never be fully automated. That is, regardless of the enormous progress achieved by transformer-based taggers, an out-domain F-score of 88% (as demonstrated by UDify on the Capitularies) falls short of the threshold that would be acceptable from the point of view of humanities scholars. And even if one assumes that F-scores around 98% are practically unattainable, since inter-rater agreements also fall short of this margin, the requirement remains for human post-processing and especially post-lemmatization, which requires corresponding lexicon-based guidance as addressed, for example, by the FLL. And since a lexicon such as the FLL requires integration with distributional semantic resources sprouting up everywhere, an answer is needed to the question of how the compact vector representations of such approaches can be mapped to manageable graphs, which in turn can be consulted by expert users to refine their lexicon work. An answer to this question will be sketched in the following section.

[25] The F1 score is the harmonic mean of precision and recall of the corresponding classification.

## 5. *Genre-sensitive Embeddings in Latin*

Gleim et al. (Gleim et al., 2019) show that using word embeddings can boost lemmatization also in Latin to a remarkable degree. However, the embeddings involved are calculated for large corpora, such as the Patrologia Latina (Jordan, 1995), without taking, for example, the genre-related diversity of text vocabularies into account: as with lexical ambiguities, such embeddings model varieties using composite structures without providing separate representations for them. Apparently, approaches of this sort assume that resources are homogeneous data: they operate on as much data as possible from genres, registers or time periods that do not exhibit substantial heterogeneity or whose heterogeneity is ignored by the model. In this article, we take a different approach: by subdividing corpora according to their contextual stratification, we obtain subcorpora for training specialized embeddings that differentiate knowledge which is otherwise amalgamated within the same model. This makes it possible to explicitly represent differences of the same word due to its varying use in different genres, subject areas or stylistic contexts. In this way, reference is made to linguistic knowledge in order to make the computation of lexical relations more transparent. The further goal is to improve the interpretability of machine learned resources from the point of view of the targeted community.

Thus, in light of Section 1, our goal is to extend FLL so that for each word a series of embeddings is learned that are differentiated according to a subset of contextual dimensions (e.g. author, genre, style, register, topic, etc.). FLL then no longer represents words (superlemmata, lemmata, wordforms or syntactic words) as nodes of a monoplex network, but as nodes of a multiplex network (Boccaletti et al., 2014) whose multiplexity is established by context dimensions. Henceforth, we denote this variant of FLL by FLL+: FLL+ is a terminological ontology that spans a multiplex network according to different contextual dimensions and thus provides a series of contextualized representations for each of its lexical entries – as (downloadable) embeddings and as traversable SemioGraphs (see below). Multiplicity means to network the vertices of the same graph according to different (in the present article: contextual, discourse-level) criteria. Furthermore, the embeddings used to network FLL+ as a multiplex network are partly hierarchically ordered by the subset relations of the corpora involved. This means, among other things, that word embeddings calculated for a subcorpus $X$ of a corpus $Y$ can be used to approximate embeddings calculated on the latter.

In order to generate FLL+, we consider genre as a contextual dimension by example of six instances (see Table 5): *epistolographic*, *legal*, *liturgical*, *narrative*, *philosophical*, *political* and *theological* texts. Further, we analyze authorship as a contextual dimension by example of three authors: Bernard of Clairvaux (d. 1153), John of Salisbury (d. 1180) and William of Ockham (d. 1347). Last but not least, we compute embeddings for our reference corpus of 111,515 texts.[26] This corpus contains the Patrologia Latina, historiographical and legal texts from the Monumenta Germaniae Historica and additional historiographical texts from the Central European Medieval Texts series. The corpus can be accessed by means of the eHumanities Desktop. The same applies to the special corpus of legal texts analyzed here (see Table 5) that contains the Corpus Iuris Civilis (compiled 528-534) and the Corpus Iuris Canonici (gradually compiled from the mid-12th to the 15th century) next to canonical decrees and Carolingian law texts. This approach of contrasting the reference corpus with specialized subcorpora makes it possible to compare embeddings generated by means of the reference with those obtained for specialized genres (see Section 6 for such a comparison).

[26] All embeddings are available for download at http://embeddings.texttechnologylab.org.

Since our focus is on genre and author-related variation and not on method optimization, we concentrate on efficiently computable methods:

1. We utilize the well-known continuous bag-of-words (CBOW) and the skip-gram model of word2vec (Mikolov et al., 2013).
2. As a further development of word2vec, we experiment with fastText (Joulin et al., 2017), which additionally evaluates character embeddings for computing word embeddings. This approach again comes in two variants: skip-gram and CBOW.
3. As we deal with subsets of corpora of varying size (see Table 5), we also experiment with a method that addresses "small" corpora. This relates to the approach of Jiang et al. (2018), who evaluate the common absence of words in text segments as an additional source of information (cf. Rieger 1989). An alternative approach is the one of Silva and Amarathunga (2019), who generate random paths in sentence networks to obtain sentence variants for extending small input corpora. Both approaches have been evaluated in the context of word similarity tasks and are promising candidates for dealing with low-resource situations. However, we concentrate on the approach of Jiang et al. (2018).

Starting from the resulting embeddings and their specialization for different genres, we generate so called *local graph views*: instead of comparing embedding representations as a whole (cf. Veremyev et al., 2019; Yaskevich et al., 2019[27]), local views generate local neighborhoods of words. For this purpose we generate all words of the FLL the graph induced by their 100 nearest neighbors. In this way, we get for each word of the FLL 50 = 10 (1 reference corpus + 9 subcorpora) × 5 (computational methods) different embeddings. Since the FLL distinguishes between wordforms, syntactic words, lemmata and superlemmata, this finally multiplies to 200 different embeddings to be managed.[28]

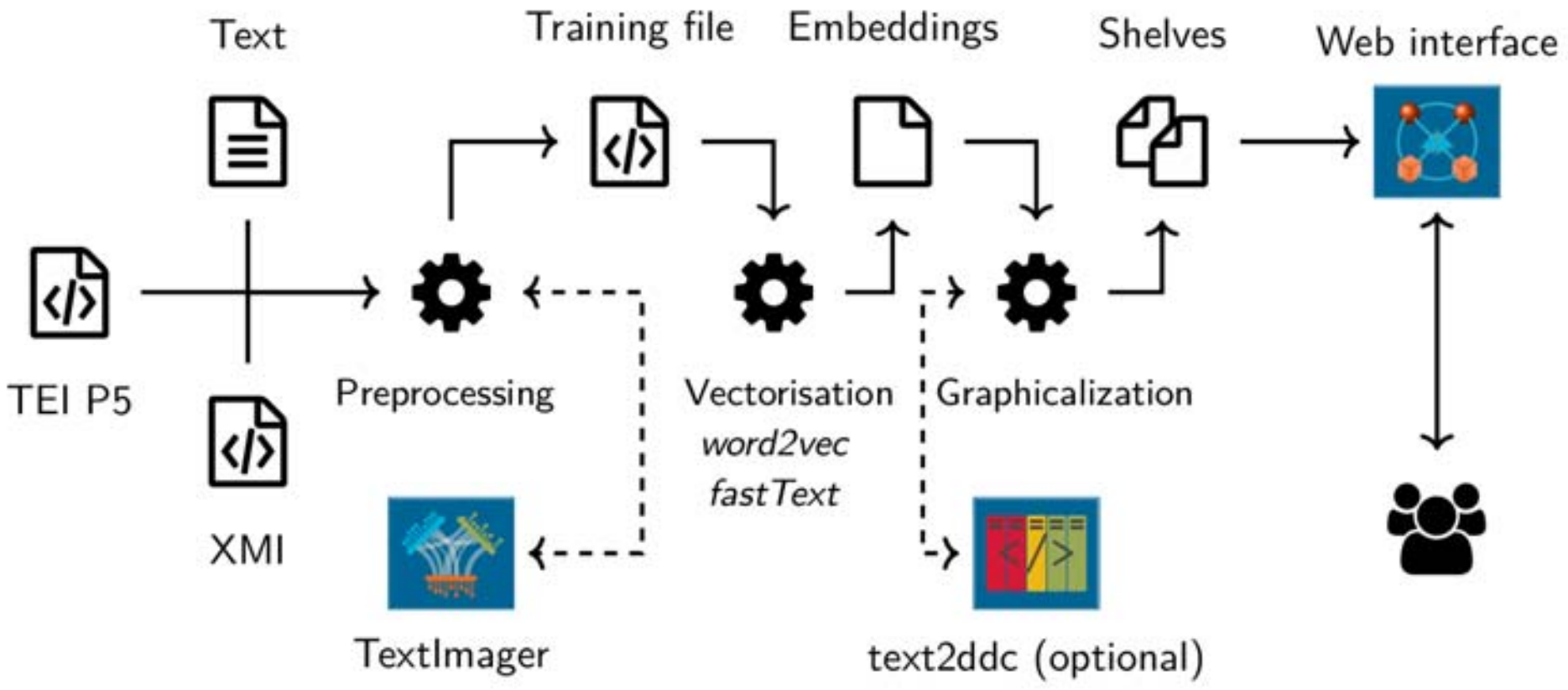

Figure 3. *Processing pipeline for generating local graph views from embeddings possibly enhanced by topic labels as input to interactive traversable SemioGraphs.*

Since our goal is not only to generate distributional semantic resources for Latin text genres, but to make them also interactively available to the (expert) user, we finally generate so-called SemioGraphs[29] as visualizations of local graph views using the pipeline[30] of Figure 3. It includes three steps to process plain text, TEI or XMI[31] documents where TextImager (Hemati et al., 2016) is used to preprocess input documents based on the procedures for processing Latin documents described in Gleim et al. (2019): preprocessing creates training files from single documents or entire repositories as input for vectorization (e.g. by means of word2vec etc.). The third step (*graphicalization*) generates input files required by the SemioGraph application, a Python script that uses the flask server framework, which is available as a Docker image for rapid setup. The SemioGraph application manages all generated embeddings for generating graph views. The output of graphicalization is stored in Python shelve files, that is, key value stores of Python objects that allow for fast access by the server. If available, graphicalization enriches SemioGraphs on the node level with topic labels using text2ddc (Uslu et al., 2019), which is currently trained for 40 languages (but not yet for Latin). The Python shelves are finally used to generate interactive visualizations using SemioGraph's web interface. In this interface, node size codes two dimensions of vertex-related salience: while *height* codes degree centrality, *width* is used to code the similarity to the seed word. Furthermore, *node transparency* can be used to code degrees of class membership values, while *node color* can map the corresponding class (this feature is not used in our example below). Beyond that, *border color* can be used to code a 2nd-level vertex-related classification (i.e. topic-related class membership). Finally, in the case of multilabel classifications, *node tiling* (i.e. pie charts) can be used to code distributions of class membership per vertex (also this feature is currently not used by our Latin SemioGraphs).

[27] See also https://github.com/anvaka/word2vec-graph.
[28] For reasons of space complexity we compute only a subset thereof.
[29] See http://semiograph.texttechnologylab.org/.
[30] This pipeline is available at https://github.com/texttechnologylab/SemioGraph.
[31] XMI is a serialization format for UIMA-CAS documents.

Generally speaking, a SemioGraph is a visual, interactive representation of word embeddings as a result of the latter procedure: starting from a word *x*, its SemioGraph displays those *m* words which by their embeddings are among the first *m* neighbors of *x* in the similarity space induced by the underlying embeddings. This means that vertices or nodes in a SemioGraph represent words or multi-word units, while edges or links represent associations of these nodes, the strength of which is represented by the thickness of the (visual representation of the) edges. Since word embeddings induce fully connected graphs (in which all words are connected with each other), the SemioGraph interface allows to filter low associations to get visual insights into the underlying graph structures: this enables the visual formation of clusters of nodes, which have a higher number of internal associations than to members of other clusters or to outliers. Thus, if a SemioGraph of a word *x* is generated using this method, this does not mean that visually disconnected words are not associated. Actually, they are, but to a degree below the user-controlled threshold value. In any case, due to the way embeddings are calculated here, SemioGraphs show paradigmatic associations. This means that even if word co-occurrences are frequent (indicating higher syntagmatic associations), the word associations need not appear in the corresponding SemioGraph. This happens in cases where the contexts in which the words are used throughout the underlying corpus are less similar than their inclusion among the *m* most similar neighbors would require: a SemioGraph always shows only a selected subset of associations. Thus, not appearing in a given SemioGraph does not necessarily mean non-existence. If the latter selection would include all words from the input corpus, then these would all be displayed in the SemioGraph, of course by means of edge representations of variable thickness. Visualizing genre-sensitive embeddings using SemioGraphs then means first generating word embeddings separately for corpora that reflect certain genre-, register-, chronology-, or other context-related features, and then visualizing the neighborhoods of certain seed words to determine the differences or similarities of their context-sensitive syntagmatic or paradigmatic associations. This is illustrated in detail in the next section by means of paradigmatic word associations.

## 6. *Brief Case Studies in Computational Historical Semantics with the Help of SemioGraph*

In this section, we apply the method of local graph views to Latin word embeddings as provided by SemioGraph, briefly present three SemioGraphs and sketch how they may provide a new kind of evidence for computational historical semantics in the humanities. In our first example we calculate paradigmatic associations (Rieger 1989) of the noun *conclusio* (*conclusion*) in the test corpus of legal texts (see Table 5). The resulting SemioGraph (see Figure 4) allows first observations on the principle func-tionality of SemioGraphs for a comparatively small corpus, on the potentials of genre-sensitive SemioGraphs, and at the same time on necessary further work and current performance of the FLL and TTLab's tagger and lemmatizer for Latin.

On the one hand, the calculated semantic connections in Figure 4 correspond to what for historians fits very well into a well-known context of legal history. First, there are mostly technical terms for different aspects in legal processes – *inquiry, examination, excuse, allegation* (*cognitio*, *examinatio*, *excusatio*, *denuntiatio*, etc.) – from the dispute to its settlement. Time expressions such as *ten-year* (*decennius*) or *four-year* (*quadriennium*) may not necessarily refer to the length of punishments, but rather to time specifications in legislation. Secondly, there is a single recognizable content, that is, marriage legislation, and this is quite clearly visible. The graph shows a vocabulary that historians would expect in texts on marriage legislation of these centuries – *divorce / repudiation* (*divortium*, *repudium*), *copulation* or *copulate* (the term *copulam* signals the need for manual post-lemmatization), the legal importance of the "consumption" of a marriage (*consummatio*), *puberty* (*pubertas*), the *conjugal union* (*matrimonialis* – adjective), or *conceiving* and *childbearing* (*partum*, specification via post-lemmatization is needed).

On the other hand, links between many words – as well as the occurrence of words without any links – indicate a need for further clarification, which must be systematized in the workflow of such queries: no link or edge connects the keyword *conclusio* with any of the other words. Apparently, c*onclusio* does not co-occur in the underlying corpus with any of these words with sufficient frequency. Links between words require a certain minimum number of neighborhoods, which serve as a reliable source of information for their paradigmatic associations. Note again that if a SemioGraph shows no link between two words, this does not mean that they are not related to each other; it only means that their paradigmatic association is below a certain minimum, where the user of the SemioGraph sets this threshold him- or herself.

Some of the key words of marriage legislation such as *copulation* or *consumption* are also disconnected in the SemioGraph in Figure 4. Some of these observations may disappear with the enlargement of the underlying corpus by means of texts that provide more evidence about their contextual similarities (Miller & Charles 1991). In any case, the calculation of paradigmatic associations ultimately aims at making such phenomena visible. That is, associations should become visible even if the words involved are rare in the underlying corpus, but the similarities of the contexts in which they are used are sufficiently strongly confirmed by that corpus. It is therefore less a matter of eliminating such observations in a SemioGraph (in terms of post-correction) than of making them (1) controllable by means of corpus selection and (2) interpretable with respect to this selection. One of several possible explanations may be that the keyword has not been used in standardized collocations: and such an observation can then be the starting point for research in the respective humanities.

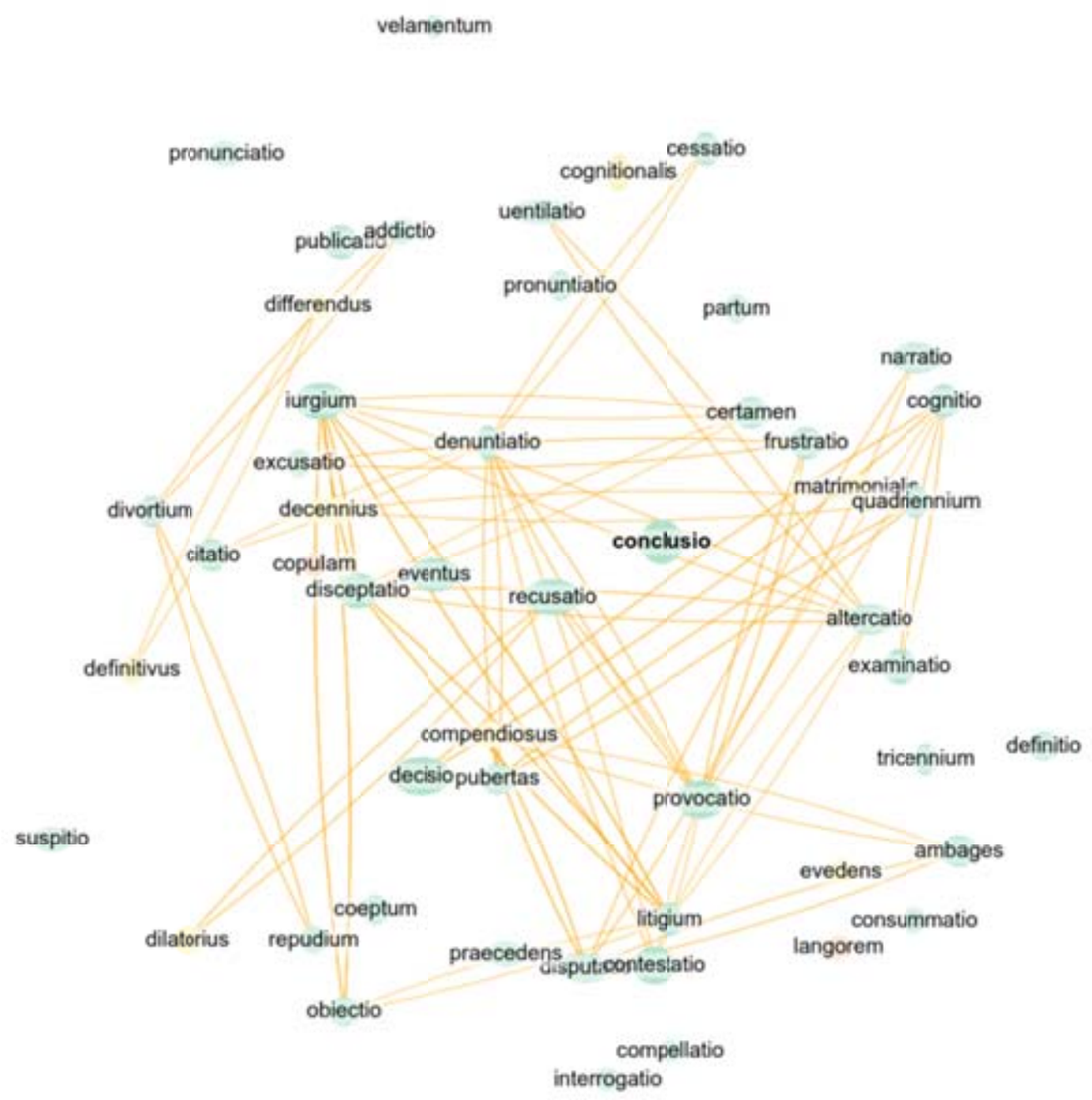

Figure 4. *Local graph view of conclusio (NN); genre: legal texts (see Table 5); method: CBOW (Mikolov et al., 2013).*

In the second example (see Figure 5) we look for the 50 closest neighbors of the verb *excommunico* ("I exclude someone from the ecclesiastical community") in our corpus of legal texts. Since excommunication was one of the few punishments the church could impose on someone, the verb can be expected mainly in legal contexts. Again, we find a semantic field that includes many tokens a historian would expect. To discuss some of them: first, *excommunico* (as well as *excommunicatio* – noun, and for post-lemmatization the spelling variations: *excomunico*, *excomunicatio*) appears in the SemioGraph together with the terms for *anatheme* (*anathema* – noun, *anathematicus* – adjective, *anathemizo* – verb, *anathematizatus* – participle) and for *interdict* (*interdictio* – noun, *interdico* – verb, *interdictus* – participle). These three legal terms are – following common encyclopedias and dictionaries such as the "*Lexikon des Mittelalters*" (Zapp, 1980, 1989, 1991) – difficult to distinguish, they were used interchangeably for describing very similar situations. Facing the current status of our genre-specific corpus of legal texts, which is still preliminary, the SemioGraph gives the impression that forms of *anathema* and of *excommunico* do indeed have a fairly similar set of neighboring words (in the sense that the SemioGraph displays many shared links). At the same time, we observe missing edges or links visualized by thin lines. This in turn indicates that the words concerned are associated below the threshold for visualizing such relations. Considering phrases like *excommunicate or anathemize* (*excommunicare vel anathematizare*) or *suspend or excommunicate* (*suspendere vel excommunicare*) used by some very influential authors (Regino of Prüm, Burchard of Worms, Ivo of Chartres), one may wonder why these co-occurrences are not reflected in the graph among the 50 closest neighbors.[32]

The same observation can be made with *interdictum*, the third central weapon for hard ecclesiastical punishment. This term shows even fewer links to *excommunico* in the SemioGraph than the forms of anatheme, although again more than 200 times *excommunico* and a form of *interdictum* co-occur in sentences of the underlying corpus. A check in the corpus shows that, although pairings such as those cited may appear sufficiently frequent overall, the individual pairings are actually not sufficiently frequent to cross the threshold. These kinds of observations lead to further questions, especially what kind of calculation – by sentences or by word distances – brings the graph closer to the notion of *paradigmatic associations*, and how the observation of paradigmatic neighborhoods relates to classical co-occurrence analyses. Other terms in the SemioGraph of Figure 5 express reasons for excommunication like *heresy* (*haeresis* – noun), *disobedience* (*inobedio* – verb), *contumacy* (*contumax* – adjective) or rebellion (*rebellis* – adjective), as well as for the lifting of the excommunication (central: *reconcilio*; among the probable terms that are according to the SemioGraph not used as neighbors of *excommunico*: *absolvo*, i.e. *absolve*). Astonishingly, we do not find expressions for the holy community of the church itself among the paradigmatic neighbors. This also would demand further investigation.

[32] One may also ask whether the SemioGraph can be associated more with methodological questions of computing rather than with historical phenomena. The reason is that any calculation of word association requires the fixation of certain parameters such as the number of neighbors in a sentence or the length of sentences in which neighbors are observed (and this holds of course also for SemioGraph). Any such parameter setting carries the risk of excluding relevant contexts – this is a general characteristic of computational linguistic analyses.

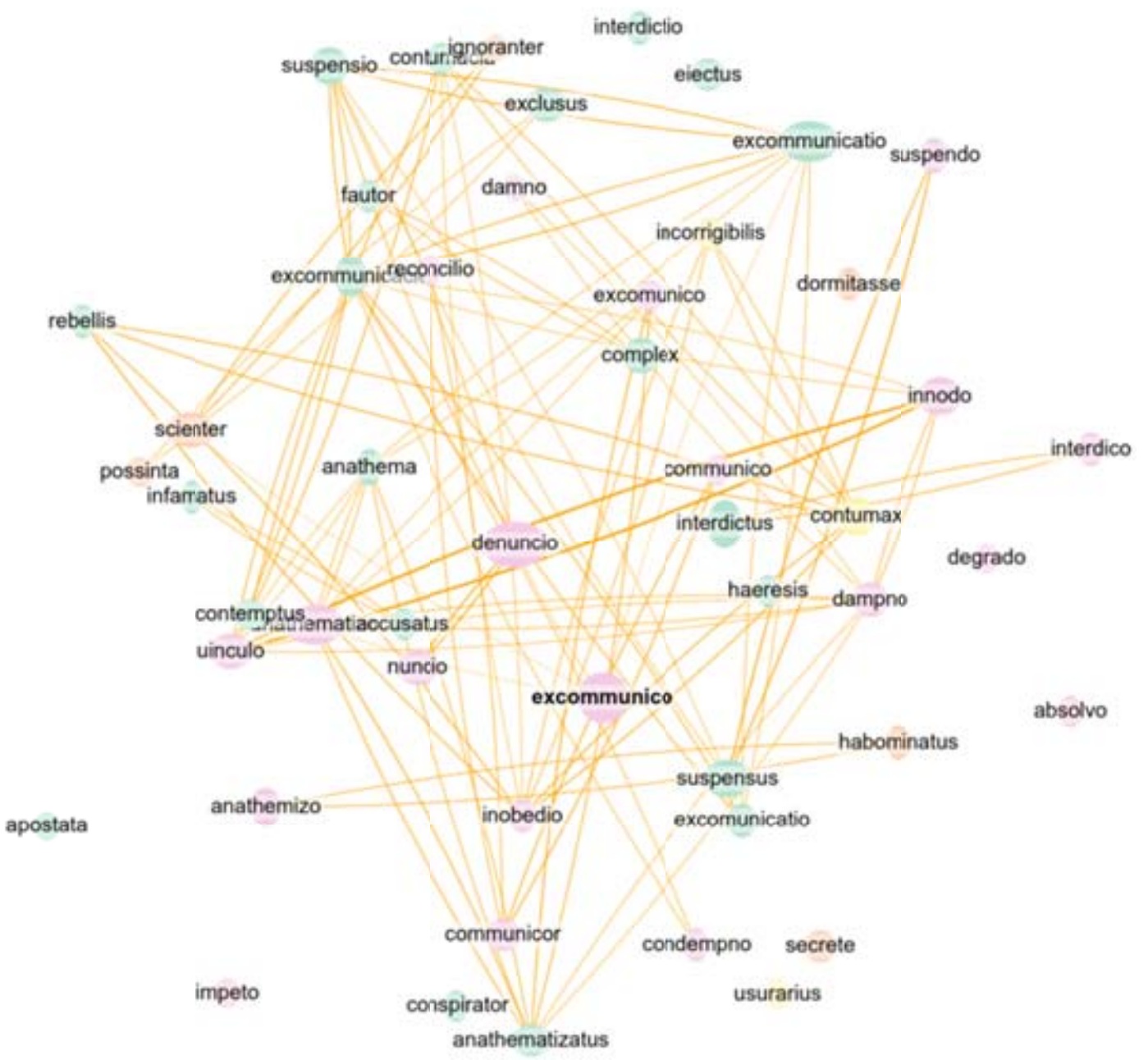

Figure 5. *Local graph view of excommunico (V); genre: legal texts; method: CBOW (Mikolov et al., 2013).*

In a third example, we look for the 50 closest associations of *father*, in Latin *pater*, first within the complete reference corpus (a mere repository from patristics to the 15th century, Figure 6) and then in our legal texts corpus (Figure 7). This experimental arrangement follows the expectation that in the corpus of all texts we might see central religious aspects (Christ as the Son of God, Mary as the Mother of God), while in the corpus of legal texts we might see the Roman Catholic normative settings of the kinship system. The SemioGraph in Figure 6, which visualizes the results for our whole repository including all sorts of texts, shows some inconspicuous, culture-unspecific words such as *father* (*pater*), *mother* (*mater*), *son* (*filius*), *uncle* (*patruus*) or *paternal* (*paternalis*). We also find words that seem to be specific to the ancient Roman Mediterranean kinship system, such as *paterfamilias*, i.e. the father as head of a household, *materfamilias* or rarely *paterfamiliae* (the word *paterfamiliae* is sometimes used; however, the wordforms that have been automatically subsumed under this lemma mostly should be subsumed under *paterfamilias*). But a first test of the diachronic distribution suggests that they are not specific to a time. This observation reminds us that the repository brings together texts from two very different social systems – from the Greco-Roman Mediterranean ("Antiquity") and from the post-Roman Latin societies ("Middle Ages"). The semantics referring to the core Christian faith are clearly recognizable. The centrality of *unigenitus* in the SemioGraph points to "God's only begotten Son" (*unigenitus dei filius*, see also *deus* and *Christus* in the SemioGraph), an often repeated phrase of the Catholic creed. The strong connection of the word *unigenitus* to consubstantiality ("of the same substance", *consubstantialis*, *consubstantialiter*) indicated by the edges in the SemioGraph stresses the link to the Catholic creed. The SemioGraph reflects the prevailing religious attitude towards paternity bonds which subordinated carnal to spiritual paternity. Fatherhood of God, priests and godparents was a strong discursive element. The important role of godparents is visible by means of the terms *co-father* and *co-mother* (*commater/conmater* and *compater/conpater*) as neighbors of *pater*. Less visible are clerics as spiritual fathers since they were simply addressed as *father*. Significant is also the lack of a broad family vocabulary that would differentiate family relationships. Only the mother and the paternal uncle (*patruus*) are present in this SemioGraph. This observation coincides at first glance with the broadly discussed hypothesis that the Roman male agnatic kinship system faded away under the influence of the church from the sixth century on in favor of kin groups organized around the conjugal couple (see Jussen, 2009). There is hardly any evidence of genealogical connections and far-reaching family relationships in this graph. This will be different in the SemioGraph in Figure 7. Odd occurrences – such as the rarely used proper name *Sopater* as part of the SemioGraph in Figure 6 – immediately raise doubts and are hence particularly important terms for cross-checks of automatic procedures, that is, for intellectual post-correction. *Sopater*, correctly lemmatized in the FFL as a proper name, is an obvious candidate to check how paradigmatic similarities and corresponding neighborhoods are calculated. Even more conspicuous is the rather central position of the term *paterfamiliae*, a very rare variant of the common but here completely missing lemma *paterfamilias*. Both are subsumed under the same superlemma within FLL. In any event, two-dimensional geometric representations of graphs should not be overinterpreted – they may be due more to the visualization method and less to the underlying graph topology.

Such obvious but rare problems, however, are contrasted in Figure 7 by a multitude of plausible and interpretatively controllable links, so that the added value of paradigmatic graphs for Latin texts can be regarded as successfully tested alongside the classical analyses of co-occurrences and syntagmatic patterns. First of all, it is striking that the kinship designations and the distinctions between the maternal and paternal lines, which were missing in the first graph, are prominent here (*avus – grandfather*, *proavus – great-grandfather*, *propatruus – great-granduncle*, *abpatruus – great-great-great uncle* from the father's side, a very rare word by the way, *tritavus – a grandfather's great-grandfather*, *proavunculus – great uncle from the mother's side*). It is also striking that the designations almost exclusively refer to the paternal line. Since canon law has developed the kinship designations in both lineages in detail (in connection with the prohibition of incest), this finding again requires verification, that is, further research by the humanities scholar. In this case, the compilation of the corpus probably needs to be corrected. Presumably, charters function linguistically differently from normative legal texts so that the corpus of legal texts should be divided into two corpora in future work. Furthermore, the data of the SemioGraph will probably only become meaningful when the long-term diachronic corpus of legal texts can be examined according to time sections. Only then will it be possible to see what was different in the Roman Mediterranean world compared to the post-Roman Latin-Christian societies.

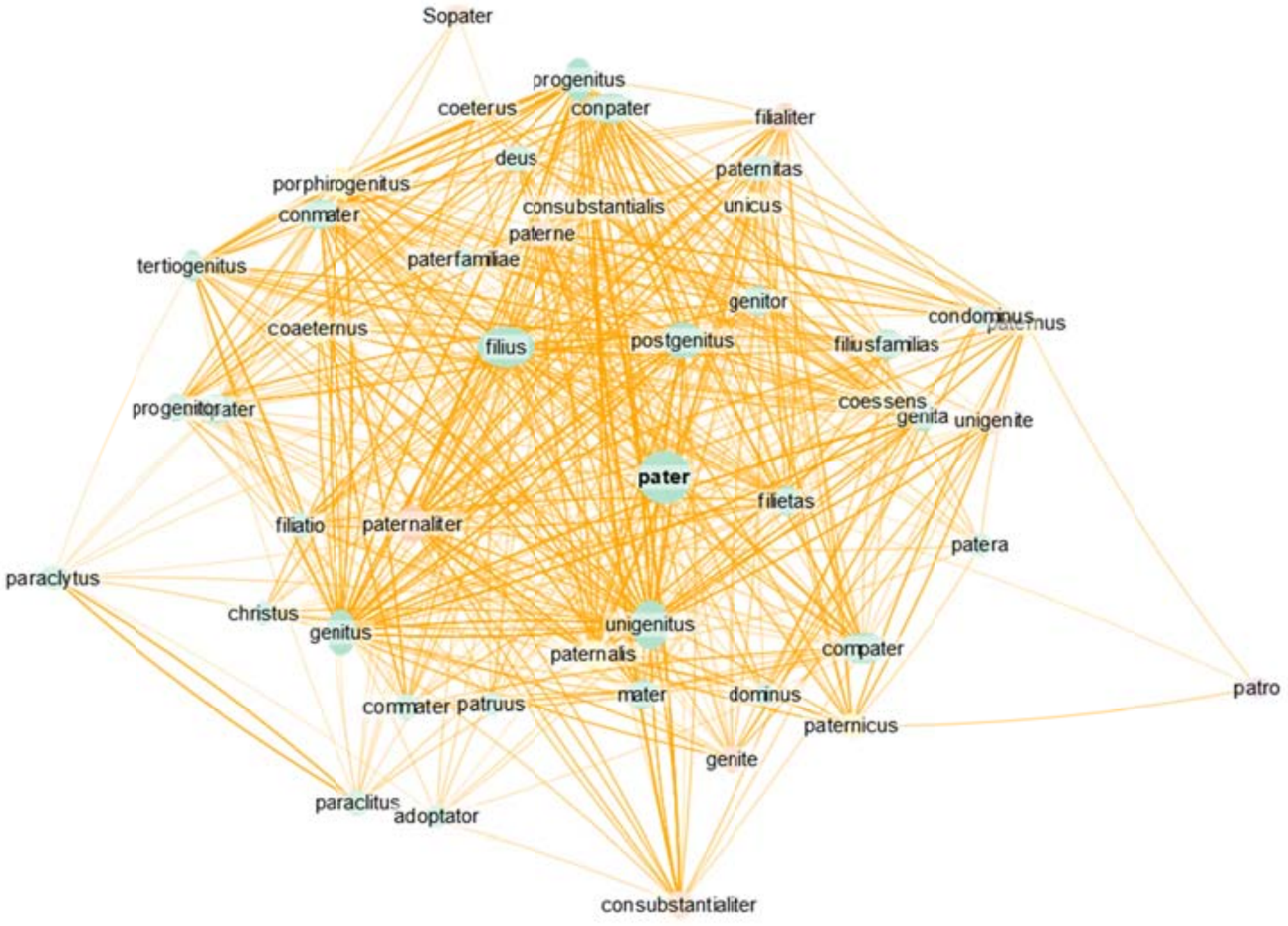

Figure 6. *Local graph view of pater (NN) taken from the reference corpus; method: CBOW (Mikolov et al., 2013).*

These short case studies may suffice as an example for the implementation of computational tools like SemioGraph and the FLL in academic cultures with a very long hermeneutical tradition:

1. The implementation of such tools in the humanities will have to start by overcoming mistrust. The results of the SemioGraphs must therefore be able to mirror the expectations and the "assured knowledge" of the humanities concerned in order to promote confidence in the reliability and controllability of the computerized calculation results.
2. Only then can they successfully manifest the unexpected that deviates from the discipline's "assured knowledge".
3. In this way, the SemioGraphs may motivate *re-reading*, no longer guided by the authority of a very long hermeneutical tradition (which inevitably privileges certain canonized 'famous' texts), but stimulated by the authority of well controllable and comparable corpora.
4. Thus, central to the acceptance of computational tools in the historical humanities is the strict and disciplined distinction between repositories and corpora in the latter sense processed by these tools.

It is these steps that must be achieved in order to institutionalize a lasting improvement of knowledge resources such as the FLL and SemioGraphs.

Establishing SemioGraphs as a tool for the visualization of paradigmatic associations in disciplines such as history or literary studies, theology or philosophy is no easy task. Despite all the changes that digitalization has brought with it, these disciplines will remain "children of hermeneutics". The success of any new methods depends on the ability to control the evidence in relatively small steps. The examples presented here can point to a way in this direction. In this article the focus was more on the technical possibilities, with some test cases as illustrations. It is left to a follow-up study to systematically verify the empirical gain – for example by examining one and the same seed word in all research perspectives mentioned here, that is, syntagmatic versus paradigmatic analyses, different definitions of neighborhoods (within one sentence, in the syntagmatic neighborhood of *n* words, etc.), comparisons of different text types and different time layers of Latin texts (Roman world up to ca. Justinian, post-Roman Latin societies 6th-11th century, 12th-16th century). Only such a multi-perspective analysis can help to assess the added value and reliability of analyses such as those exemplified here.

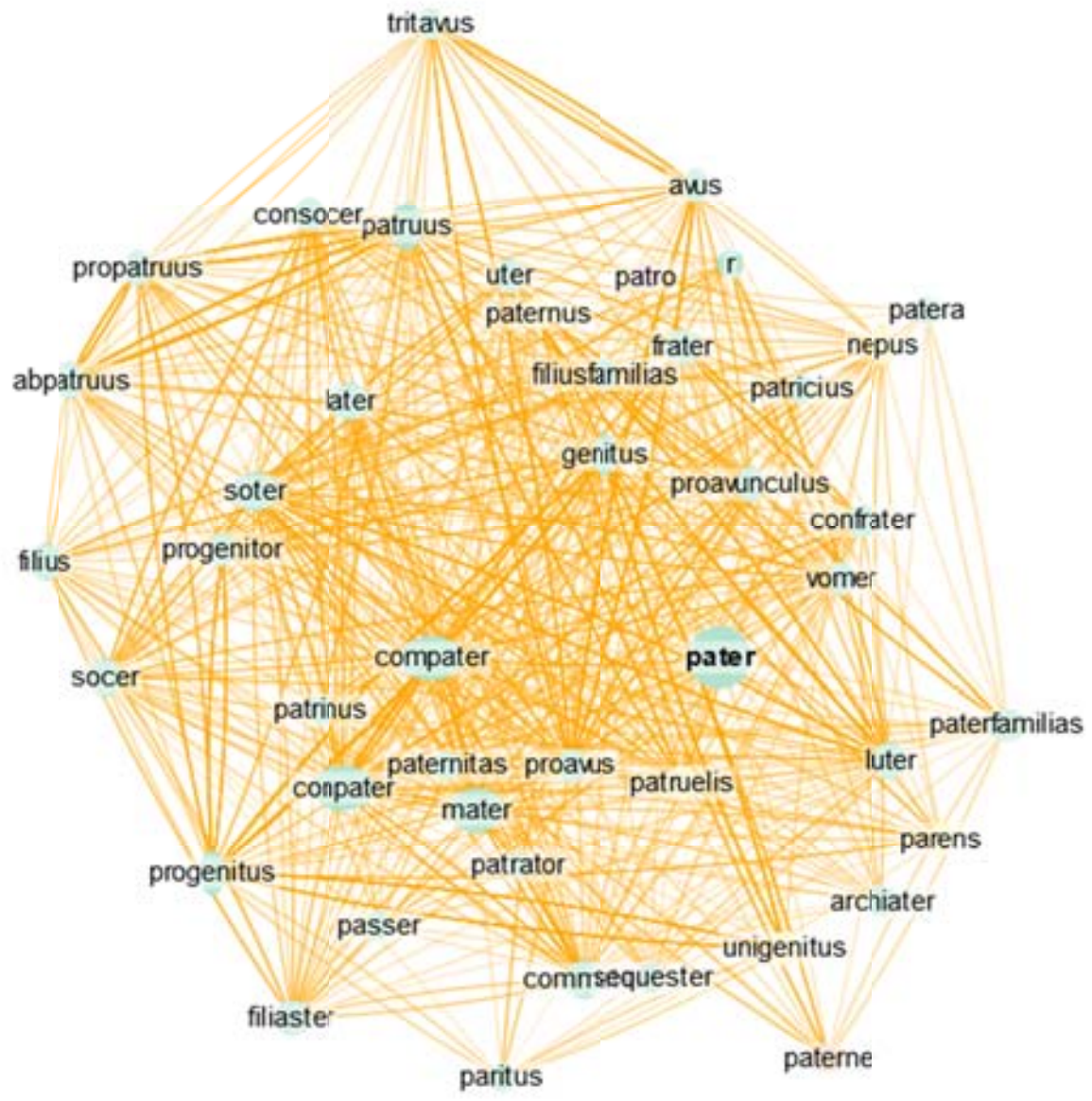

Figure 7. *Local graph view of pater (NN); genre: legal texts; method: CBOW (Mikolov et al., 2013).*

By means of these case studies, we obtain an example of the triadic role of computational tools such as SemioGraph from the perspective of the applying humanities. That is, such tools serve

(1) to meet and confirm the expectations of the scholars involved,
(2) to manifest the unexpected that deviates from the current state of knowledge of the discipline, and
(3) to motivate subsequent processes of *re-reading* in order to substantiate possible interpretations of the unexpected finding.

As far as this "new reading" is equipped with tool chains of the kind outlined in this article, it could eventually lead to updates of the underlying knowledge resources, that is, the FLL and the embeddings based thereon, which in turn require updates of corresponding SemioGraphs, so that we finally get a manifestation of a digitally enhanced hermeneutic circle. We are convinced that it is worth pursuing this research direction further.

## 7. *Conclusion*

In this article, we presented the *Frankfurt Latin Lexicon* (FLL) as a dictionary resource for Latin that distinguishes between word forms, syntactic words, lemmata and superlemmata and thus implements a word model known from the Wiktionary project. We outlined a restricted crowdsourcing process by means of which the FLL is continuously checked and updated as well as the lemmatization of texts based thereon. We additionally reported progress in the lemmatization of Latin texts and stressed the need to enhance the FLL by means of word embeddings that are stratified according to contextual parameters such as genre, authorship and chronological order. Then, we introduced SemioGraphs as a means to interact with and traverse this embedding information. Finally, we presented three case studies based on SemioGraphs using word embeddings computed for selected seed words of the FLL.

Since these case studies show the need for downstream processes of close reading and possibly for corrections of the underlying lemmatization, we have identified in this process chain an instance for a *digitally enhanced hermeneutic circle*. It could be seen as an example of a prototypical strategy for dealing with lemmatization or, more generally, natural language processing of historical language texts. Future work will focus on a more detailed examination of word embeddings in Latin, their local and global graph representations, and in particular on their intrinsic evaluation.

### *Acknowledgement*

We thank our anonymous reviewers for their valuable comments and critical suggestions.

### *References*

BAMMAN, D. and CRANE, G. (2011), *The ancient Greek and Latin dependency treebanks*, in: Language technology for cultural heritage, pp. 79–98. Berlin: Springer.

BOCCALETTI, S., BIANCONI, G., CRIADO, R., DEL GENIO, C. I., GÓMEZ-GARDENES, J., ROMANCE, M. SENDINA-NADAL, I., WANG, Z. and ZANIN, M. (2014), *The structure and dynamics of multilayer networks*, in «Physics Reports» 544(1), 1–122.

CECCHINI, F. M., PASSAROTTI, M., MARONGIU, P. and ZEMAN, D. (2018), *Challenges in converting the Index Thomisticus treebank into Universal Dependencies*, in *Proceedings of the Universal Dependencies Workshop 2018 (UDW 2018)*, Brussels.

CLARK, K., KHANDELWAL, U., LEVY, O. and MANNING, C. D. (2019), *What does BERT look at? An analysis of BERT's attention*, in arXiv preprint arXiv:1906.04341.

CRANE, G. (1996), *Building a digital library: the Perseus project as a case study in the humanities*, in *Proceedings of the first ACM international conference on Digital libraries, DL '96*, New York, pp. 3–10, ACM.

DEVLIN, J., CHANG, M.-W., LEE, K. and TOUTANOVA, K. (2019), *BERT: Pre-training of deep bidirectional transformers for language understanding*, in *Proceedings of the 2019 Conference of the North American Chapter of the Association for Computational Linguistics: Human Language Technologies*, Volume 1, Minneapolis, Minnesota, pp. 4171–4186.

EGER, S., GLEIM, R. and MEHLER, A. (2016), *Lemmatization and Morphological Tagging in German and Latin: A comparison and a survey of the state-of-the-art*, in LREC 2016.

FRANZINI, G., PEVERELLI, A., RUFFOLO, P., PASSAROTTI, M., SANNA, H., SIGNORONI, E., VENTURA, V. and ZAMPEDRI, F. (2019), *Nunc Est Aestimandum. towards an evaluation of the Latin WordNet*, in CLiC-it 2019.

GLEIM, R., EGER, S., MEHLER, A., USLU, T., HEMATI, W., LÜCKING, A., HENLEIN, A., KAHLSDORF, S. and HOENEN, A. (2019), *A practitioner's view: a survey and comparison of lemmatization and morphological tagging in German and Latin*, in «Journal of Language Modeling», DOI: http://dx.doi.org/10.15398/jlm.v7i1.205

GLEIM, R., MEHLER, A. and Ernst, A. (2012), *SOA implementation of the eHumanities Desktop*, in *Proceedings of the Workshop on Service-oriented Architectures (SOAs) for the Humanities: Solutions and Impacts, Digital Humanities*, Hamburg.

HALLIDAY, M. A. K. and Hasan, R. (1976), *Cohesion in English*. London: Longman.

HAUG, D. T. and JØHNDAL, M. (2008), *Creating a parallel treebank of the old Indo-European bible translations*, in *Proceedings of the Second Workshop on Language Technology for Cultural Heritage Data (LaTeCH 2008)*, pp. 27–34.

HEMATI, W., USLU, T. and MEHLER, A. (2016), TextImager: a distributed UIMA-based system for NLP, in *Proc. of COLING 2016: System Demonstrations*, pp. 59–63.

JAKOBSON, R. (1971), *Selected Writings II. Word and Language*, The Hague: Mouton.

JIANG, C., YU, H.-F., HSIEH, C.-J. and CHANG, K.-W. (2018), *Learning word embeddings for low-resource languages by PU learning*, in *Proceedings of the 2018 Conference of the North American Chapter of the Association for Computational Linguistics: Human Language Technologies*, Volume 1 (Long Papers), New Orleans, pp. 1024–1034.

JORDAN, M. D. (Ed.) (1995), *Patrologia Latina database*, Cambridge: Chadwyck-Healey.

JOULIN, A., GRAVE, E., BOJANOWSKI, P. and MIKOLOV, T. (2017), *Bag of tricks for efficient text classification*, in *Proceedings of the 15th Conference of the EACL: Volume 2*, Short Papers, Valencia, Spain, pp. 427–431.

JURŠIC, M., MOZETIC, I., ERJAVEC, T. and LAVRAC, N. (2010), *LemmaGen: Multilingual lemmatisation with induced ripple-down rules*, in «Journal of Universal Computer Science» 16(9), pp. 1190–1214.

JUSSEN, B., MEHLER, A. and ERNST, A. (2007), *A corpus management system for historical semantics*, in «Sprache und Datenverarbeitung. International Journal for Language Data Processing» 31(1-2) (2007): pp. 81-89.

JUSSEN, B. (2009), *Perspektiven der Verwandtschaftsforschung fünfundzwanzig Jahre nach Jack Goodys »Entwicklung von Ehe und Familie in Europa«*, in SPIEß, K.-H. (2009, ed.), *Die Familie in der Gesellschaft des Mittelalters, Vorträge und Forschungen*, pp. 275–324. Ostfildern: Thorbecke.

KOMNINOS, A. and MANANDHAR, S. (2016), *Dependency based embeddings for sentence classification tasks*, in *Proceedings of the 2016 Conference of the North American Chapter of the Association for Computational Linguistics: Human Language Technologies*, pp. 1490–1500.

KONDRATYUK, D., GAVENCIAK, T., STRAKA, M. and HAJIC, J. (2018), *LemmaTag: Jointly tagging and lemmatizing for morphologically-rich languages with BRNNs*, in CoRR abs/1808.03703.

KONDRATYUK, D. and STRAKA, M. (2019), *75 languages, 1 model: Parsing universal dependencies universally*, in *Proceedings of the 2019 Conference on Empirical Methods in Natural Language Processing and the 9th International Joint Conference on Natural Language Processing (EMNLP-IJCNLP)*, Hong Kong, China, pp. 2779–2795.

KOSTER, C. H. A. and Verbruggen, E. (2002), *The AGFL Grammar Work Lab*, in *Proceedings of FREENIX/Usenix*, pp. 13–18.

LEVY, O. and GOLDBERG, Y. (2014), *Dependency-based word embeddings*, in *Proceedings of the 52nd Annual Meeting of the Association for Computational Linguistics*, Volume 2, pp. 302–308.

LING, W., DYER, C., BLACK, A. and TRANCOSO, I. (2015), *Two/Too Simple Adaptations of word2vec for Syntax Problems*, in *Proceedings of the 2015 Conference of the North American Chapter of the Association for Computational Linguistics: Human Language Technologies*.

MEHLER, A., DIEWALD, N., WALTINGER, U., GLEIM, R., ESCH, D., JOB, B., KÜCHELMANN, T., PUSTYLNIKOV, O. and BLANCHARD, P. (2011), *Evolution of Romance language in written communication: Network analysis of late Latin and early Romance corpora*, in «Leonardo» 44(3), pp. 244–245.

MEHLER, A., GLEIM, R., HEMATI, W. and USLU, T. (2017), *Skalenfreie online soziale Lexika am Beispiel von Wiktionary*, in ENGELBERG, S., LOBIN, H., STEYER, K. and WOLFER, S. (2017, eds.), *Proceedings of 53rd Annual Conference of the Institut für Deutsche Sprache (IDS)*, March 14-16, Mannheim, pp. 269–291. Berlin: De Gruyter.

MEHLER, A., VOR DER BRÜCK, T., GLEIM, R. and GEELHAAR, T. (2015), *Towards a network model of the coreness of texts: An experiment in classifying Latin texts using the TTLab Latin Tagger*, in Biemann, C. and Mehler, A. (2015, eds.), *Text Mining: From Ontology Learning to Automated Text Processing Applications, Theory and Applications of Natural Language Processing*, pp. 87–112. Berlin/New York: Springer.

MENGE, H. (2009), *Lehrbuch der lateinischen Syntax und Semantik* (4. Aufl., unveränd. Nachdr. der 2., überarb. Aufl., ed.), Darmstadt: Wissenschaftliche Buchgesellschaft.

MIKOLOV, T., CHEN, K., CORRADO, G. and DEAN, J. (2013), *Efficient estimation of word representations in vector space*, in arXiv preprint arXiv:1301.3781.

MIKOLOV, T., W. YIH, and G. ZWEIG (2013), Linguistic regularities in continuous space word representations, in *Proceedings of NAACL 2013*, pp. 746–751.

MILLER, G. A. (1995), *WordNet: A lexical database for English*, in Commun. ACM 38(11), 39–41.

MILLER, G. A. and Charles, W. G. (1991), *Contextual correlates of semantic similarity*, in Language and Cognitive Processes 6(1), pp. 1–28.

MINOZZI, STEFANO (2017), Latin WordNet, una rete di conoscenza semantica per il latino e alcune ipotesi di utilizzo nel campo dell'Information Retrieval. *Antichità 14.* doi.org/10.14277/6969-182-9/ANT-14-10.

MÜLLER, T., SCHMID, H. AND SCHÜTZE, H. (2013), *Efficient higher-order CRFs for morphological tagging*, in *Proceedings of the 2013 Conference on Empirical Methods in Natural Language Processing*, pp. 322–332. ACL.

NIVRE, J., DE MARNEFFE, M.-C., GINTER, F., GOLDBERG, Y., HAJIC, J., MANNING, C. D., MCDONALD, R., PETROV, S., PYYSALO, S., SILVEIRA, N., TSARFATY, R. and ZEMAN, D. (2016), *Universal Dependencies v1: A multilingual tree-bank collection*, in *Proceedings of the Tenth International Conference on Language Resources and Evaluation (LREC'16)*, pp. 1659–1666.

PASSAROTTI, M. (2004), *Development and perspectives of the Latin morphological analyser LemLat*, in «Linguistica Computazionale» 20–21.

PASSAROTTI, M. and DELL'ORLETTA, F. (2010), *Improvements in parsing the Index Thomisticus Treebank. Revision, combination and a feature model for medieval Latin*, in *Proceedings of LREC 2010*. ELDA.

RAIBLE, W. (1981), *Von der Allgegenwart des Gegensinns (und einiger anderer Relationen), Strategien zur Einordnung semantischer Informationen*, in «Zeitschrift für romanische Philologie» 97 (1-2), pp. 1–40.

RIEGER, B. B. (1989), *Unscharfe Semantik: die empirische Analyse, quantitative Beschreibung, formale Repräsentation und prozedurale Modellierung vager Wortbedeutungen in Texten*, Frankfurt am Main: Lang.

RUBENBAUER, H., J., HOFMANN, B. and HEINE, R. (2009), *Lateinische Grammatik* (12., korr. Aufl. ed.), Bamberg: Oldenbourg.

SAUSSURE, FERDINAND DE (1916), *Cours de linguistique générale*. Edited by C. Bally and A. Sechehaye. Lausanne/Paris: Payot.

SCHADT, H. (1982), *Die Darstellungen der Arbores consanguinitatis und der Arbores affinitatis: Bildschemata in juristischen Handschriften: Teilw. zugl.: Tübingen, Univ., Diss.*, 1973, Tübingen: Wasmuth.

SCHMID, H. (1994), *Probabilistic part-of-speech tagging using decision trees*, in JONES D. and SOMERS H. (1994, eds.), *New Methods in Language Processing Studies in Computational Linguistics*, UCL Press.

SHOEYBI, M., PATWARY, M., PURI, R., LEGRESLEY, P., CASPER, J. and CATANZARO, B. (2019), *Megatron-LM: Training multi-billion parameter language models using GPU model parallelism*, in arXiv preprint arXiv:1909.08053.

SILVA, A. and AMARATHUNGA, C. (2019), *On learning word embeddings from linguistically augmented text corpora*, in *Proceedings of the 13th International Conference on Computational Semantics* – Short Papers, Gothenburg, pp. 52–58. ACL

SOWA, J. F. (2000), *Knowledge Representation: Logical, Philosophical, and Computational Foundations*, Brooks/Cole.

STOECKEL, M., HENLEIN, A., HEMATI, W. and MEHLER, A. (2020), Voting for POS tagging of Latin texts: Using the flair of FLAIR to better Ensemble Classifiers by Example of Latin, in *Proceedings of LT4HALA 2020 – 1st Workshop on Language Technologies for Historical and Ancient Languages*, Marseille, France, European Language Resources Association (ELRA), 130-135, https://www.aclweb.org/anthology/2020.lt4hala-1.2

STRAKA, M. and STRAKOVÁ, J. (2017), *Tokenizing, POS tagging, lemmatizing and parsing UD 2.0 with UDPipe*, in *Proceedings of the CoNLL 2017 Shared Task: Multilingual Parsing from Raw Text to Universal Dependencies*, Vancouver, Canada, pp. 88–99. ACL.

USLU, T., MEHLER, A. and BAUMARTZ, D. (2019), *Computing Classifier-based Embeddings with the Help of text2ddc*, in *Proceedings of the 20th International Conference on Computational Linguistics and Intelligent Text Processing, (CICLing 2019).*

VEREMYEV, A., SEMENOV, A., PASILIAO, E. L., and BOGINSKI, V. (2019), *Graph-based exploration and clustering analysis of semantic spaces*, in «Applied Network Science» 4(1), pp. 104.

VOR DER BRÜCK, T. and MEHLER , A. (2016), *TLT-CRF: A lexicon-supported morphological tagger for Latin based on conditional random fields*, in *LREC 2016*.

YASKEVICH, A., LISITSINA, A., KATRICHEVA, N., ZHORDANIYA, T., KUTUZOV, A. and KUZMENKO, E. (2019), *vec2graph: a Python library for visualizing word embeddings as graphs*, in *Proceedings of AIST*, Kazan.

ZAPP, H. (1980): Art. Anathem, in: Lexikon des Mittelalters. Vol. 1, p. 574.

ZAPP, H. (1989): Art. Exkommunikation, in: Lexikon des Mittelalters. Vol. 4, p. 170.

ZAPP, H. (1991): Art. Interdikt, in Lexikon des Mittelalters. Vol. 5, p. 466–467.